\documentclass[a4paper]{article}

 \PassOptionsToPackage{numbers, compress}{natbib}
\pdfoutput=1
\usepackage{hyperref}
\hypersetup{
 pdfinfo={
   Title={Mapping Informal Settlements in Developing Countries with Multi-resolution, Multi-spectral Data},
   Author={Patrick Helber, Bradley Gram-Hansen, Indhu Varatharajan, Faiza Azam, Alejandro Coca-Castro, Veronika Kopackova, Piotr Bilinski}
 }
}
\usepackage[final]{pdfpages}
\usepackage{wrapfig}
\usepackage[preprint]{nips_2018}

\usepackage[utf8]{inputenc} % allow utf-8 input
\usepackage[T1]{fontenc}    % use 8-bit T1 fonts
\usepackage{hyperref}       % hyperlinks
\usepackage{url}            % simple URL typesetting
\usepackage{booktabs}       % professional-quality tables
\usepackage{amsfonts}       % blackboard math symbols
\usepackage{nicefrac}       % compact symbols for 1/2, etc.
\usepackage{microtype}      % microtypography
\usepackage{graphicx}
\usepackage[polish,english]{babel}

\title{Mapping Informal Settlements in Developing Countries with Multi-resolution, Multi-spectral Data}
\author{Patrick Helber\thanks{authors contributed equally} \\
     DFKI and TU Kaiserlautern \\
     \texttt{\href{mailto:patrick.helber@dfki.de}{\small{patrick.helber@dfki.de}}} \\
    \And 
   Bradley Gram-Hansen{$^*$}\\
     University of Oxford \\
     \texttt{\href{mailto:bradley@robots.ox.ac.uk}{\small{bradley@robots.ox.ac.uk}}}\\
    \And
    Indhu Varatharajan \\
     DLR \\
     \texttt{\href{mailto:indhu.varatharajan@dlr.de}{\small{indhu.varatharajan@dlr.de}}}\\
    \And
    Faiza Azam \\
     Independent researcher \\
     \texttt{\href{mailto:fazam@hotmail.de}{\small{fazam@hotmail.de}}}\\
    \And 
    Alejandro Coca-Castro \\
     King's College London \\
     \texttt{\href{mailto:alejandro.coca\textunderscore castro@kcl.ac.uk}{\small{alejandro.coca\_castro@kcl.ac.uk}}}\\
         \And 
    Veronika Kopackova \\
     Czech Geological Survey \\
     \texttt{\href{mailto:veronika.kopackova@seznam.cz}{\small{veronika.kopackova@seznam.cz}}} \\
         \And 
    Piotr Biliński \\
     University of Oxford and University of Warsaw \\ % and University of Warsaw, Poland \\
     \texttt{\href{mailto:piotrb@robots.ox.ac.uk}{\small{piotrb@robots.ox.ac.uk}}}\\
     }

\begin{document}

\maketitle

\begin{abstract}
 Detecting and mapping informal settlements encompasses several of the United Nations sustainable development goals. This is because informal settlements are home to the most socially and economically vulnerable people on the planet. Thus, understanding where these settlements are is of paramount importance to both government and non-government organizations (NGOs), such as the United Nations Children’s Fund (UNICEF), who can use this information to deliver effective social and economic aid. We propose two effective methods for detecting and mapping the locations of informal settlements.
 One uses only low-resolution (LR), freely available, Sentinel-2 multispectral satellite imagery with noisy annotations, whilst the other is a deep learning approach that uses only costly very-high-resolution (VHR) satellite imagery. To our knowledge, we are the first to map informal settlements successfully with low-resolution satellite imagery. We extensively evaluate and compare the proposed methods. Please find additional material at \mbox{\url{https://frontierdevelopmentlab.github.io/informal-settlements/}}.
 
 %!! This  is  in  contrast  to  previous  studies  that  onlyuse costly very-high resolution (VHR) satellite and aerial im-agery.
 
 % However, data regarding informal and formal settlements is primarily unavailable and if available is often incomplete. This is due to the cost and complexity in gathering it on a large scale. Moreover, the definition of \emph{informal settlements} is also very broad, which makes it a non-trivial task to collect data and thus teach a machine what to look for. Due to these limitations we openly provide two datasets with both high and low resolution imagery from a variety of locations, with the accompanying annotations, enabling other machine learning practitioners to compare methodologies.Then

 % In addition to this, we show how we can detect informal settlements by combining both domain knowledge and machine learning techniques,  to build a classifier that looks for known roofing materials used in informal settlements.
 
\end{abstract}

\section{Introduction}
%\begin{figure}[t!]
%\centering
%\includegraphics[width=0.5\textwidth, height=3cm, %keepaspectratio]{images/kibera_divide.jpg}
%\caption{Image of the divide between formal and informal settlements in %Kibera, Nairobi. \textit{ Permission granted by Johnny Miller and Unequal %Scenes}.}
%\end{figure}

% The United Nations~(UN) and the Organisation for Economic Co-operation and Development~(OECD) state that informal settlements are defined as 
The United Nations~(UN) state that informal settlements are defined as
follows~\cite{2008oecd,united2012state}: 
\begin{quote}
\begin{enumerate}
\item Inhabitants have no security of tenure vis-\`a-vis the land or
dwellings they inhabit, with modalities ranging from squatting to informal rental housing. 
\item The neighborhoods usually lack, or are cut off from, basic services and city infrastructure.
\item The housing may not comply with current planning and building regulations, and is often situated in geographically and environmentally hazardous areas.
\end{enumerate}
\end{quote}

%Furthermore, informal settlements can be a form of real estate speculation for all income levels of urban residents, affluent and poor. 

Slums are the most deprived and excluded form of informal settlements. They can be characterized by poverty and large agglomerations of dilapidated housing, located in the most hazardous urban land, near industries and dump sites, in swamps, degraded soils and flood-prone zones~\cite{Kohli2016}. In addition to tenure insecurity, slum dwellers lack basic infrastructure and services, public space and green areas, and are constantly exposed to eviction, disease and violence~\cite{Sclar2005}. The ability to map and locate these settlements would give organizations such as UNICEF and other related organizations the ability to provide effective social and economic aid~\cite{pais2002poverty}.

% Typically, those that live in informal settlements are the most vulnerable in society, subject to harsh social and economic constraints~\cite{WEKESA2011238}. Although informal settlements are well studied in the humanities and remote sensing communities \cite{fincher2003planning,WEKESA2011238,united2012state,huchzermeyer2006informal,hofmann2008detecting}, in machine learning little has been done to map informal settlements using high-resolution satellite imagery~\cite{mahabir2018critical,mboga2017detection,varshney2015targeting} and nothing has been done using low-resolution imagery.   Being able to map and locate these settlements would give organizations such as UNICEF and other UN organizations the ability to provide effective social and economic aid~\cite{pais2002poverty}. This in turn would enable those communities to evolve in a sustainable way, allowing the people living in those environments to gain a much better quality of life. Moreover, in tackling this problem we are directly addressing multiple of the UN sustainable development goals~\cite{UNsus}, which aim to eliminate poverty, increase good health and well-being, provide quality education, clean water and sanitation, affordable and clean energy, sustainable work and economic growth, access to industry, innovation and infrastructure.\\ 

\begin{figure}[t!]
\centering
  \includegraphics[width=0.72\linewidth]{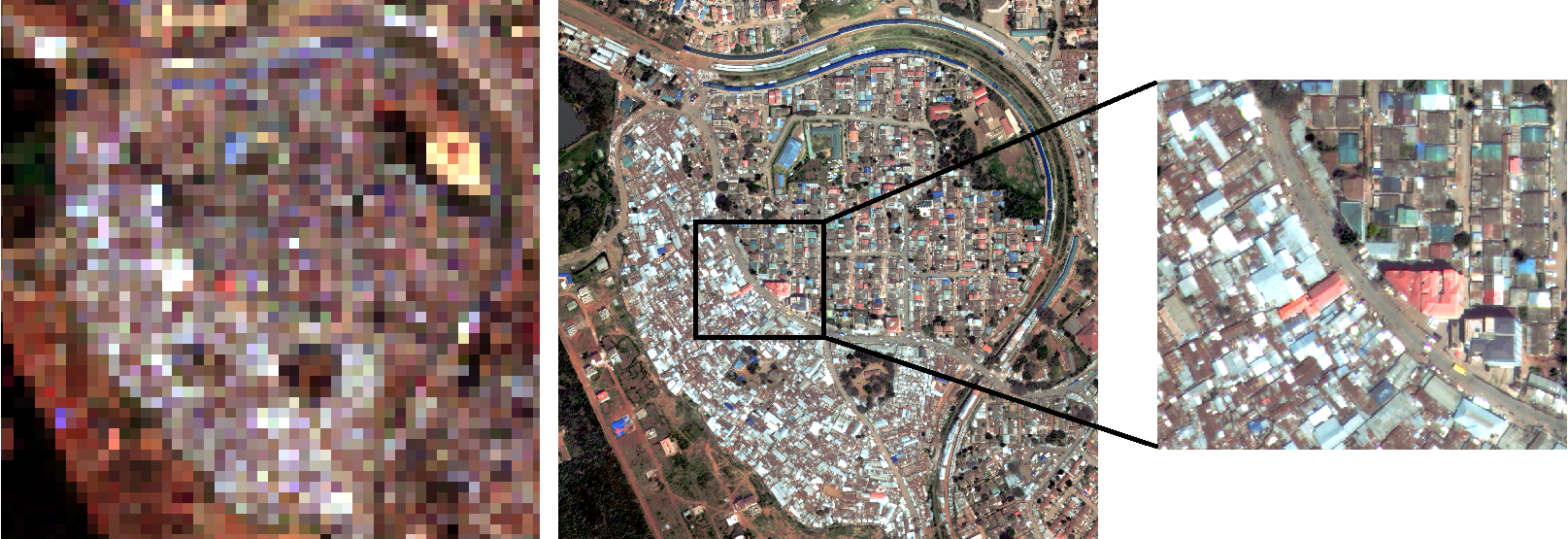}

% \raggedleft
%\begin{minipage}{.32\textwidth}
%  \includegraphics[width=0.70\linewidth]{images/kibera_2.png}
%   \caption{figure1}
%   \label{fig:test1}
%\end{minipage}%
%\begin{minipage}{.32\textwidth}
%  \includegraphics[width=0.70\linewidth]{images/kibera_1.png}
%   \caption{figure}
%   \label{fig:test2}
%\end{minipage}
%\begin{minipage}{.32\textwidth}
%  \includegraphics[width=0.70\linewidth]{images/kibera_slum_excerpt_full_resolution.png}
%   \caption{figure}
%   \label{fig:test2}
%\end{minipage}

\caption{Two images of the same informal settlement in Kibera, Nairobi representing the difference between high and low resolution imagery. \textit{Left}: The Sentinel-2 10m resolution image. \textit{Right}: A DigitalGlobe 30cm very-high-resolution image. Also, a detailed view of a part of the very-high-resolution image is shown.}
\label{fig:compresimg}
\end{figure}

\subsection{Data}
In this paper, we have annotated low-resolution (LR) and very-high-resolution (VHR) satellite imagery for the locations of informal settlements in parts of Kenya, South Africa, Nigeria, Sudan, Colombia and Mumbai. 

\label{sec:data}

\paragraph{Sentinel-2 Multi-spectral Satellite Data}

The Sentinel-2 mission is part of the Copernicus programme, a global earth observation service. The Sentinel-2 satellites map the entire global land mass on average every 5 days at various resolutions of 10 to 60$m$ $per$ $pixel$. Sentinel-2 provides multiple-resolution, multi-spectral imagery using a multi-spectral instrument~(MSI). MSI measures top of the atmosphere radiances in 13 spectral bands covering the visible, near infrared and the shortwave infrared part of the electromagnetic spectrum at different spatial resolution depending upon the particular band~\cite{sent2userhandbook,Zhang2017}. We only use 10 bands due to atmospheric distortions. The resolution of up to 10$m$ denotes that each pixel represents a $10m\times10m$ surface, which means that there is a certain amount of contextual information contained within one pixel. By observing the spectral signal, which provides us with the chemical composition of a pixel, we can extract this contextual information.

\paragraph{Very-High-Resolution Satellite Images}
In addition to freely available multi-spectral low-resolution satellite images, we use very-high-resolution images with a resolution of up to 30$cm$ per pixel, kindly provided by DigitalGlobe through the Satellite Applications Catapult. See Figure~\ref{fig:compresimg} to view the difference in the resolution between Sentinel-2 and very-high-resolution imagery. 

\section{Methods}
\label{sec:method}

% In this section, we describe our approaches for  detecting and mapping informal settlements. We introduce two different methodologies. 
Our \textbf{first method} trains a classifier to learn what the spectral signal of an informal settlement is, using low-resolution freely available spectral data. To do this, we employ a pixel-wise classification, where the classifier learns whether or not a 10-band spectra is associated to an informal settlement or the environment, which encompasses everything that is not an informal settlement. 

When we require finer grained features, such as the roof size, or the density of the surrounding settlements to determine whether or not there exists an informal settlement, we demonstrate our \textbf{second method}, which trains a state-of-the-art semantic segmentation deep neural network that uses very-high-resolution satellite imagery. This is crucial when informal settlements do not have unique spectra when compared to the environment, like those in Al Geneina, Sudan, see Figure~\ref{fig:sudaninffor}.

\paragraph{Pixel-wise Classification with Canonical Correlation Forest} Canonical Correlation Forests~(CCFs)~\cite{rainforth2015canonical} are a decision tree ensemble method for classification and regression.
CCFs are the state-of-the-art random forest technique, which have shown to achieve remarkable results for numerous regression and classification tasks~\cite{rainforth2015canonical}.
Individual canonical correlation trees are binary decision trees with hyperplane splits based on local canonical correlation coefficients calculated during training. Like most random forest based approaches, CCFs have very few hyper-parameters to tune and typically provide very good performance out of the box. The only parameter that has to be set is the number of trees, $n_{trees}$. For CCFs, setting $n_{trees} = 15$ provides a performance that is empirically equivalent to a random forest that has $n_{trees} = 500$~\cite{rainforth2015canonical}, meaning CCFs have lower computational costs, whilst providing better classification.
CCFs work by using canonical correlation analysis~(CCA) and projection bootstrapping during the training of each tree, which projects the data into a space that maximally correlates the inputs with the outputs. This is particularly useful when we have small datasets, as for our case, because it reduces the amount of artificial randomness required to be added during the tree training procedure and improves the ensemble predictive performance~\cite{rainforth2015canonical}.\\

 The computational efficiency aspects of CCFs and their suitability to both small and large datasets, makes them ideal for detecting informal settlements as many of the organisations that we aim to help will not have access to a large amount of compute resources, therefore computational efficiency is important. 
%  in addition to this, to run the CCFs for both training and prediction, all that has to be called is one function. This ensures that the end user does not need to be an expert in ensemble methods and makes the method akin to plug and play. This means that we have to use the data as efficiently as possible, which CCFs allow us to do. 
If the access to very-high-resolution imagery and the computational costs are not a restriction, we can employ a deep learning approach using convolution neural networks~(CNN) to detect informal settlements.
% Although one maybe able to use a convolution neural network~(CNN), we have chosen not to for the following reasons.  Each pixel in a Sentinel-2 image approximately represents a  $10m\times10m$ region.  Therefore, the spectral information of each pixel contains a lot of contextual information. As the spectral information provides us with information regarding the composition of materials within an image, this include the types of vegetation, soils, metals, bricks and any other materials. However, a CCN on one pixel, or multiple pixels, would not be able to determine this level of contextual information.

% Finally, the computational cost in training a CNN, in terms of time and monetary cost, is typically expensive, which would prohibit many of the organizations that we aim to help.
% \textbf{One of the challenges with this approach is that it is difficult to validate the predictions of our models robustly, as we do not have access to accurate data regarding the materials that each pixel represents. However, by comparing what we know to be a representation of the ground truth, the model seemingly provides meaningful predictions, but is difficult to validate in a robust way.\\
\paragraph{Contextual Classification with CNNs}

Since informal settlements can also be classified by the rooftop size and the surrounding building density, we employ a state-of-the-art semantic segmentation neural network on optical~(RGB) very-high-resolution satellite imagery to detect these contextual features. Contextual features are important when it is not possible to distinguish informal settlements from the environment by spectral signal in the same region. An example of such an informal settlement is shown in Figure~\ref{fig:sudaninffor}. We see that the informal settlements in a rural region of Al Geneina, Sudan have a very low building density, and also the roof tops of both formal and informal settlements are built out of concrete, meaning they have the same spectral signal. This is in contrast to the dense slums in Nairobi and Mumbai.

% \begin{figure}[t!]
% \includegraphics[width=0.45\textwidth]{images/kibera_dl_ccf.png}
% \caption{}
% \label{fig:compDLandCCF}
% \end{figure} 
 \setcounter{figure}{2}    
\begin{wrapfigure}{r}{0.45\textwidth}
\vspace*{-0.4cm}
\centering
%\begin{minipage}{.5\textwidth}
%\raggedleft
%  \includegraphics[width=0.55\linewidth]{images/sudan_1_a.png}
%   \caption{figure1}
%   \label{fig:test1}
%\end{minipage}%
%\begin{minipage}{.5\textwidth}
%\raggedright
  \includegraphics[width=1.0\linewidth]{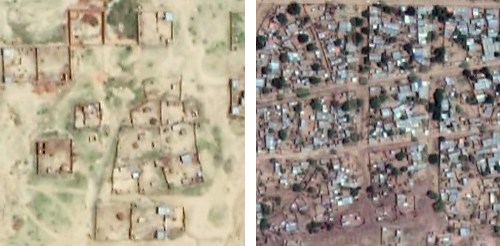}
%   \caption{figure}
%   \label{fig:test2}
%\end{minipage}
\caption{Very-high-resolution images provided by DigitalGlobe comparing an informal, \textit{left} and formal settlement, \textit{right}, in Al Geneina, Sudan. The main distinguishing feature is the wider contextual information, as the material spectrums are the same.}
\label{fig:sudaninffor}
\end{wrapfigure}
 % deliberately odd figure placement
 \setcounter{figure}{1}    
\begin{figure*}[t!]
\centering
\begin{minipage}[b]{0.325\textwidth}
   \includegraphics[height=1.9cm, keepaspectratio]{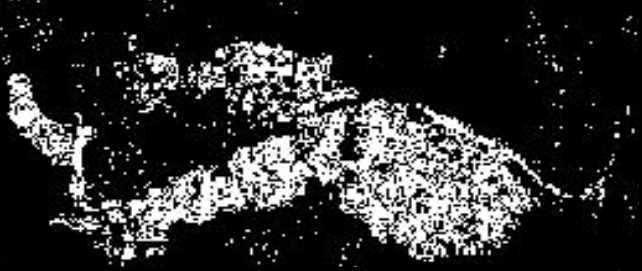}
%    \caption{The CCF prediction of informal settlements in Kibera on LR Sentinel-2 spectral imagery}
%    \label{fig:Ng2} 
\end{minipage}
\begin{minipage}[b]{0.325\textwidth}
   \includegraphics[height=1.9cm, keepaspectratio]{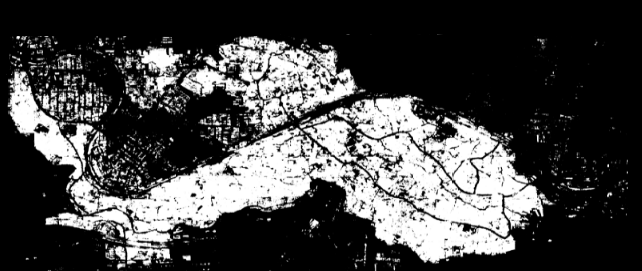}
%    \caption{DeepLab V3 + prediction of informal settlements in Kibera, trained on VHR imagery}
%    \label{fig:Ng1} 
\end{minipage}
\begin{minipage}[b]{0.325\textwidth}
   \includegraphics[height=1.9cm, keepaspectratio]{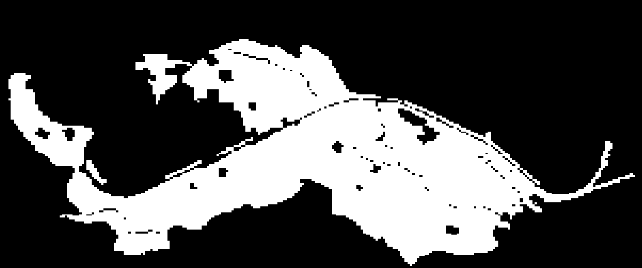}
%    \caption{The ground truth informal settlement mask for Kibera}
%    \label{fig:Ng3}
\end{minipage}

\caption{\small{Predictions of informal settlements (white pixels) in Kibera, Nairobi. \textit{Left:} The CCF prediction of informal settlements in Kibera on low-resolution Sentinel-2 spectral imagery. \textit{Middle:} CNN based prediction of informal settlements in Kibera, trained on very-high-resolution imagery. \textit{Right: }The ground truth informal settlement mask for Kibera.}}
\label{fig:compDLandCCF}
\end{figure*}

\paragraph{Encoder-Decoder with Atrous Separable Convolution}
We use the DeepLabv3+ encoder-decoder architecture. DeepLabv3+ ~\cite{deeplabv3plus2018} is a deep CNN that extends the prior DeepLabv3 network with a decoder module to refine the segmentation results of the previous encoder-decoder architecture particularly at the object boarders. The DeepLab architecture uses Atrous Spatial Pyramid Pooling~(ASPP) with atrous convolutions to explicitly control the resolution at which feature responses are computed within the CNN. 
This ASPP module is augmented with image level features to capture longer range information. 
We use a Xception 65 network backbone in the encoder-decoder architecture. The beneficial use of this Xception model together with applying depth wise separable convolution to ASPP and the decoder modules have been shown in~\cite{deeplabv3plus2018}. 

%\subsubsection{Implementation details}

%We train the entire network end-to-end with the usual back-propagation algorithm using
%eight Tesla V100 GPUs with 16 GBs of memory each.
%We initialize the layer weights using those from the pre-trained PASCAL VOC 2012 model~\cite{pascal-voc-2012}. We then fine-tune in turn
%the finer strides on the training/validation data.
%We train our deep network with a batch size of 32, an initial learning rate of 0.001 and a learning rate decay factor of 0.1 every 2.000 steps until convergence. Our experiments are based on a single-scale evaluation. All other hyper-parameters are the same as in the
%DeepLabv3+ model~\cite{deeplabv3plus2018}.

%\subsection{Material Pixel-wise Classification with CCFs}

%Our third approach to map informal settlements consists of creating a classifier using a CCF for detecting the spectral signals of different types of materials. These materials are known to be used in the construction of informal settlements~\cite{kuffer2016slums}. In contrary, the first approach consists of building a classifier to find a common spectral signal for informal settlements. In Africa alone, Afrobarometer reports that nearly 75\% of roofs on informal settlements are made out of metals such as tin or aluminum, with the remaining 25\% built out of thatch, plastic, concrete or asbestos~\cite{Afro}. 

\section{Results}
\label{sec:results}
% \begin{figure}[h!]
% \centering
% \includegraphics[width=0.45\textwidth]{images/aoi.png}
% \caption{A global map representing the locations of our case studies.}
% \label{fig:mapofall}
% \end{figure}
% Veronika comments : I think we should explain more that VHR is not a solution to do global mapping...
%moreover, VHR would not give us info about the roof material, which is crucial for instance in case of asbestos....
 
% We have ensured that the images and annotations are aligned in space and time to reduce any additional noise in the data.

\paragraph{Experimental Setup} For each region we have a 10$m$ low-resolution Sentinel-2 image, the corresponding very-high-resolution 30-50$cm$ resolution image and the ground truth annotations. When training and validating a model on the same region we use a 80-20 split. We ensure that each class contains the same number of points, we then randomly sample 80\% of each class to generate the training data and then use the remaining 20\% of each class to construct our test set, which is comprised of a different set of points. We then center the training data (testing data accordingly) to have a mean of zero and standard deviation of one. We set the $n_{trees} =10$ for training the CCF.
For validating our methods we report both pixel accuracy, and mean intersection over union~(mIoU). 

%We use the standard definition of mean IOU, $ meanIOU = \frac{1}{n_{class}}\frac{t_{ii}}{(t_{i} + \sum_j n_{ji} - n_{ii})} $ and pixel classification, $pxclass = \frac{\sum_i n_{ii}}{t_i}$, where $n_{class}$ is the total number of classes, $n_{ij}$ is the number of pixels of class $i$ predicted to belong to class $j$, and $t_i$ is the total number of pixels of class $i$ in ground truth segmentation.\\

We provide a comparison of both the pixel-wise classification with CCFs and the contextual classification with CNNs for the detection and mapping of informal settlements, see on the left side of Table~\ref{tab:res3}. The CCFs trained solely on freely available and easily accessible low-resolution data perform well, although they are unable to match the performance of the CNN trained on very-high-resolution imagery, except for Kibera. Figure~\ref{fig:compDLandCCF} shows the predictions of both methods and the ground truth annotations.  Despite having access to very-high-resolution data, the CNN still manages to miss-classify structural elements of the informal settlements in Kibera. Whereas the CCF, although more granular, incorporates the full structure of the informal settlement in Kibera via only the spectral information.

\paragraph{Generalizability} To demonstrate the adaptability of our approach we train each model on different parts of the world and use that model to perform predictions on other unseen regions across the globe. For this paper we train two models, one on Northern Nairobi, Kenya and another on Medellin, Colombia. The results can be found in the right-half of Table~\ref{tab:res3}. Even though we only have a small amount of data, we are able to demonstrate that our models can generalize moderately well, even with data that is noisy and partially incomplete. This opens a new challenge for transfer learning.

% We provide several more results in the supplementary materials.
 
% \begin{table}
% \begin{tabular}{|l|l|l|l|}
% \hline
% & Pixel Classification & IOU & Mean IOU \\ \hline
% DeepLab &  &  &  \\ \hline
% CCFs & 68\% & 62\% & 54\% \\ \hline
% \end{tabular}
% \caption{Results for informal settlement detection pixel classification, informal and mean IOU, when trained on a different region.}
% \label{tab:res1}
% \end{table}

\begin{table}[h!]\small
\centering
\caption{\small{\textit{Left-half}: Pixel accuracy and mean IOU (\%) results for informal settlement detection using the CCF pixel-wise classification and the contextual classification with CNNs. CCFs are trained and tested on low-resolution imagery, CNNs are trained and tested on very-high-resolution imagery. \textit{Right-half}: Using the CCFs we train a model on Northern Nairobi~(NN) and a model trained on Medellin~(M), then predict on all regions. *ground truth annotations are less than 75\% complete for the region.}}
% to add statement about LR and VHR
\noindent\hrulefill

\smallskip\noindent
\resizebox{\linewidth}{!}{%
\begin{tabular}{lcc|cc||cc|cc}
 & \multicolumn{2}{c}{\textbf{Pixel Acc.}} & \multicolumn{2}{c}{\textbf{Mean IOU}} & \multicolumn{2}{c}{\textbf{Pixel Acc.}} & \multicolumn{2}{c}{\textbf{Mean IOU}} \\ \midrule
\textbf{Region} & \textbf{CCF(LR)} & \textbf{CNN(VHR)} & \textbf{CCF(LR)} & \textbf{CNN(VHR)} & \textbf{NN} & \textbf{M} & \textbf{NN} & \textbf{M} \\ \midrule
Kenya, Northern Nairobi & 69.4 & 93.1 & 62.0 & 80.8 & 69.4 & 55.0 & 62.0 & 54.4 \\
Kenya, Kibera & 69.0 & 78.2 & 73.3 & 65.5 & 67.3  & 63.8 & 54.1 & 56.0\\
South Africa, Capetown* & 92.0 & - & 33.2 & - & 41.3 & 71.5 & 43.1 & 32.0 \\
Sudan, El Daien & 78.0 & 86.0 & 61.3 & 73.4 & 14.2 & 1.1 & 37.9 & 34.0 \\
Sudan, Al Geneina & 83.2 & 89.2 & 35.7 & 76.3 &  27.1 & 6.0 & 34.9 & 41.0\\
Nigeria, Makoko* & 76.2 & 87.4 & 59.9 & 74.0 & 59.0 & 77.0 & 37.8 & 34.6 \\
Colombia, Medellin* & 84.2 & 95.3 & 74.0 & 83.0 &  65.0 & 84.2 & 46.9 & 74.0 \\
India, Mumbai* & 97.0 & - & 40.3 & - &  37.9 & 63.0 & 32.4 & 34.4
\end{tabular}}
\label{tab:res3}
\end{table}

% \section{Conclusions}% and Future Work}

\paragraph{Conclusion}

In this work we have
% composed a series of annotated ground truth datasets and have 
% provided for the first time benchmarks for detecting informal settlements
provided benchmarks for detecting informal settlements
% We have provided a comprehensive list of the challenges faced in mapping informal settlements and some of the constraints faced by NGOs. 
and have proposed two different methods for detecting informal settlements. The first method uses computationally efficient CCFs to learn the spectral signal of informal settlements from multispectral low-resolution satellite imagery. The second trains a CNN with very-high-resolution satellite imagery to extract finer grained features. We extensively evaluated the proposed methods and demonstrated the generalization capabilities of the computationally efficient CCFs to detect informal settlements globally. In particular, to the best of our knowledge, we demonstrated for the first time that informal settlements can be detected effectively using only freely and openly accessible multi-spectral low-resolution satellite imagery.

\paragraph*{Acknowledgments.}
This project was executed during the Frontier Development Lab~(FDL), Europe program, a partnership between the Phi-Lab at ESA, the Satellite Applications~(SA) Catapult, Nvidia Corporation, Oxford University and Kellogg College. We gratefully acknowledge the support of Adrien Muller and Tom Jones of SA Catapult for their useful comments, providing very-high-resolution imagery and ground truth annotations for Nairobi. We thank UNICEF, in particular Do-Hyung Kim and Clara Palau Montava, for valuable discussions and AIData for access to geo-located Afrobarometer data. We thank Nvidia for computation resources. We thank Yarin Gal for his helpful comments. Patrick Helber was supported by the NVIDIA AI Lab program and the BMBF project DeFuseNN (Grant 01IW17002). Bradley Gram-Hansen was also supported by the UK EPSRC CDT in Autonomous Intelligent Machines and Systems.

\small
\bibliographystyle{plain}
\bibliography{refs}
%\includepdf[pages=1-last]{main.pdf}
\end{document}